%% file: Main.tex
\documentclass[10pt,journal,compsoc]{IEEEtran}
\ifCLASSOPTIONcompsoc
  \usepackage[nocompress]{cite}
\else
  \usepackage{cite}
\fi

\ifCLASSINFOpdf
\else
\fi
\usepackage{algorithm} 
\usepackage{algpseudocode}
\algnewcommand{\Initialize}[1]{%
  \State \textbf{Initialize:}
  \Statex \hspace*{\algorithmicindent}\parbox[t]{.8\linewidth}{\raggedright #1}
}
\usepackage{multirow}
\usepackage{tabularx,booktabs}
\usepackage{amsmath,amsfonts,amssymb}
\DeclareMathOperator*{\argmax}{arg\,max}

\usepackage{subfig}
\usepackage{adjustbox}
\usepackage{xcolor}
\usepackage{comment}

\AtBeginDocument{%
\providecommand\BibTeX{{%
    \normalfont B\kern-0.5em{\scshape i\kern-0.25em b}\kern-0.8em\TeX}}}
  
\usepackage{url}
\usepackage{hyperref}
\usepackage{lipsum}

\begin{document}

\title{Learning Data Teaching Strategies Via Knowledge Tracing}

\author{Ghodai Abdelrahman,
        Qing Wang
\IEEEcompsocitemizethanks{\IEEEcompsocthanksitem All authors are with Research School of Computer Science, 
 Australian National University, Canberra,
ACT, 0200.\protect\\

E-mail: {ghodai.abdelrahman, 
qing.wang@anu.edu.au}}
}

%
%

\markboth{Research Article}%
{Shell \MakeLowercase{\textit{et al.}}: Bare Demo of IEEEtran.cls for Computer Society Journals}

\IEEEtitleabstractindextext{%
\begin{abstract}
Teaching plays a fundamental role in human learning. Typically, a human teaching strategy would involve assessing a student's knowledge progress for tailoring the teaching materials in a way that enhances the learning progress. A human teacher would achieve this by tracing a student's knowledge over important learning concepts in a task. Albeit, such teaching strategy is not well exploited yet in machine learning as current machine teaching methods tend to directly assess the progress on individual training samples without paying attention to the underlying learning concepts in a learning task. In this paper, we propose a novel method, called \emph{Knowledge Augmented Data Teaching} (KADT), which can optimize a data teaching strategy for a student model by tracing its knowledge progress over multiple learning concepts in a learning task. Specifically, the KADT method incorporates a knowledge tracing model to dynamically capture the knowledge progress of a student model in terms of latent learning concepts. Then we develop an attention pooling mechanism to distill knowledge representations of a student model with respect to class labels, which enables to develop a data teaching strategy on critical training samples. We have evaluated the performance of the KADT method on four different machine learning tasks including knowledge tracing, sentiment analysis, movie recommendation, and image classification. The results comparing to the state-of-the-art methods empirically validate that KADT consistently outperforms others on all tasks.

\end{abstract}

}

\maketitle

\IEEEdisplaynontitleabstractindextext

%
\IEEEpeerreviewmaketitle

\input{Introduction.tex}

\input{Problem}
\input{Methodology}

\input{Experiments}

\input{Results}

\input{RelatedWork}

\input{Conclusion.tex}

\bibliographystyle{IEEEtran}
\bibliography{References}

\end{document}

%% file: Introduction.tex
\section{Introduction}
\label{sec:introduction}

The ability to digest knowledge has always been a vital characteristic of human intelligence. It is known that a student's learning performance is not only determined by her ability to digest different learning concepts but also significantly affected by the teaching strategy of her teacher. A good teacher would optimize a teaching strategy of learning materials, exercises, and problem-solving techniques to enable a student to achieve her learning objectives. This is typically performed by tracing a student's knowledge progress over important learning concepts in a learning task, e.g., addition, subtraction, and multiplication in elementary math course. A human teacher can evolve such a teaching strategy according to the performance level of a student. This developmental evolution of a teaching strategy is the key to unlocking students' full potential at different levels.



A question arising is: \emph{can a machine learn to teach in a similar manner to a human teacher?} In a machine learning scenario, a teaching strategy often includes the task of prioritizing training data (equivalent to learning materials in human learning), the choice of a loss function (equivalent to assessments in human learning), and the hyper-parameter configuration of a hypothesis function (equivalent to problem-solving techniques in human learning). A machine may target one or more of these teaching dimensions to evolve an effective teaching strategy.


In the past years, a number of attempts were made to optimize the training procedure of a machine learning student model. Curriculum learning methods~\cite{CL2009,SpitkovskyAJ10,Graves17} aimed at ranking training samples based on their difficulty levels to build a good learning curriculum for a student model. Similarly, self-paced learning (SPL) methods~\cite{KumarPK10,LeeG11,JianSPL14} used a hardness threshold that gradually increases with the progress of a student to build a training data curriculum. Machine teaching methods~\cite{liu2017iterative,liu2016teaching} focused on selecting optimal training samples that minimize a teaching cost (e.g., the size of a training set). Dynamic loss functions~\cite{L2TDF18} and graduated optimization~\cite{HazanLS16} methods adjust the difficulty of a loss function according to the progress of a student's learning. Despite considerable progress being made, these methods either depend on heuristic rules (e.g., hardness or difficulty thresholds) or assume a pre-defined student model to drive a teaching strategy. 

Recently, reinforcement learning (RL) has been proposed to develop teaching strategies~\cite{L2T,L2TDF18}. Generally, it involves two building blocks: a teacher model and a student model. A teacher model aims to optimize a teaching strategy, while a student model follows the teaching strategy to optimize its learning objective. 
However, these existing works have several limitations. First, they depend on hand-crafted states, ignoring the fact that a student model may have different performances on different learning concepts in a learning task. Second, they require a careful assignment of a target performance threshold for each learning task based on sparse reward functions over a state space (e.g., positively rewarding a teacher model only if a student model performs above a specified threshold value). This demands task-specific expertise during the RL training to land on an effective teaching policy~\cite{rl2018}. 

\begin{figure}[t!]
\includegraphics[width=8.65cm, height=5.7cm]{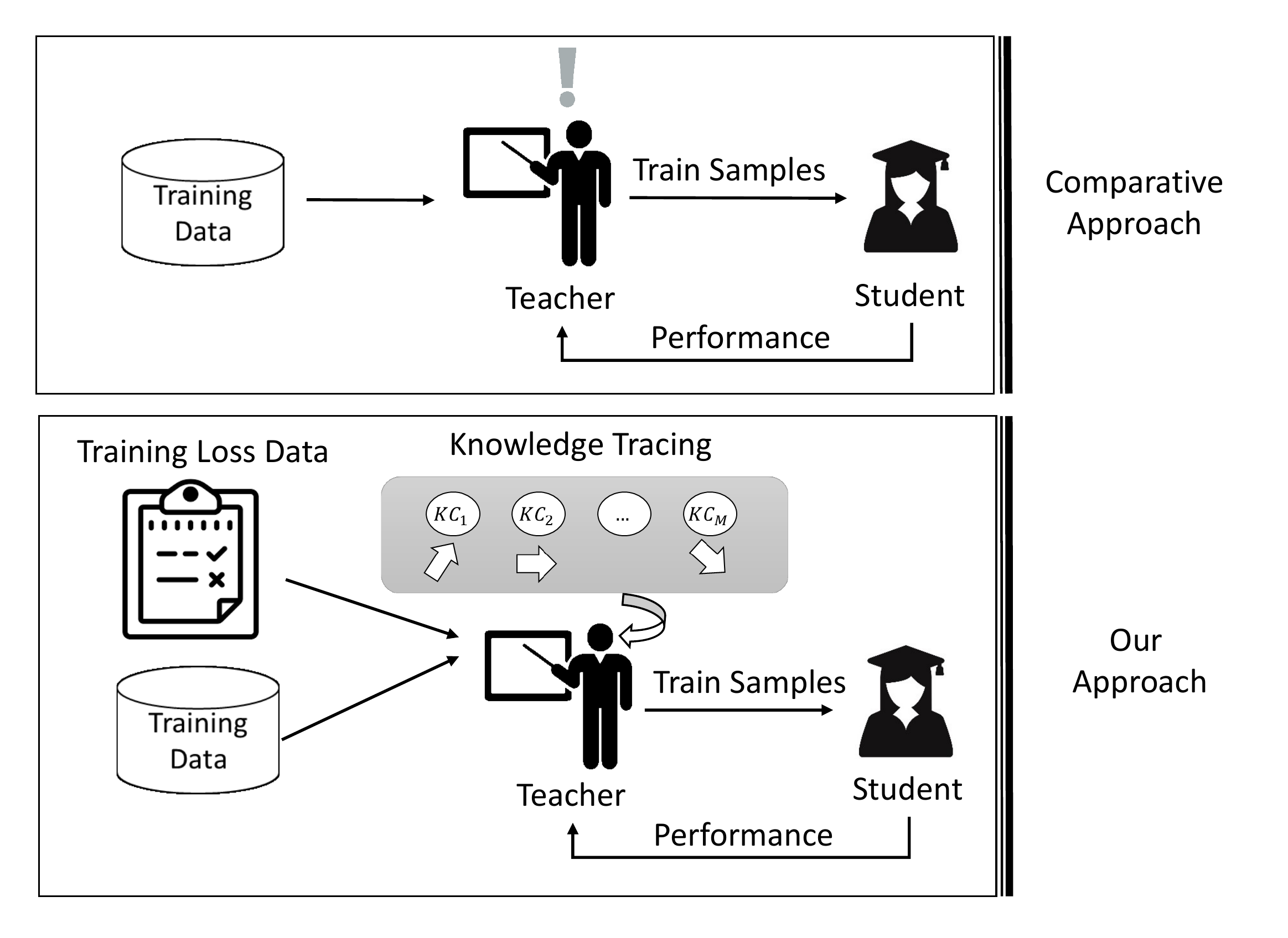}\vspace{-0.4cm}
\caption{Comparing the proposed KADT method and a conventional data teaching method.}\vspace{-0.2cm}
\label{fig:money}
\end{figure}

\begin{figure*}[t!]
\includegraphics[width=0.95\linewidth]{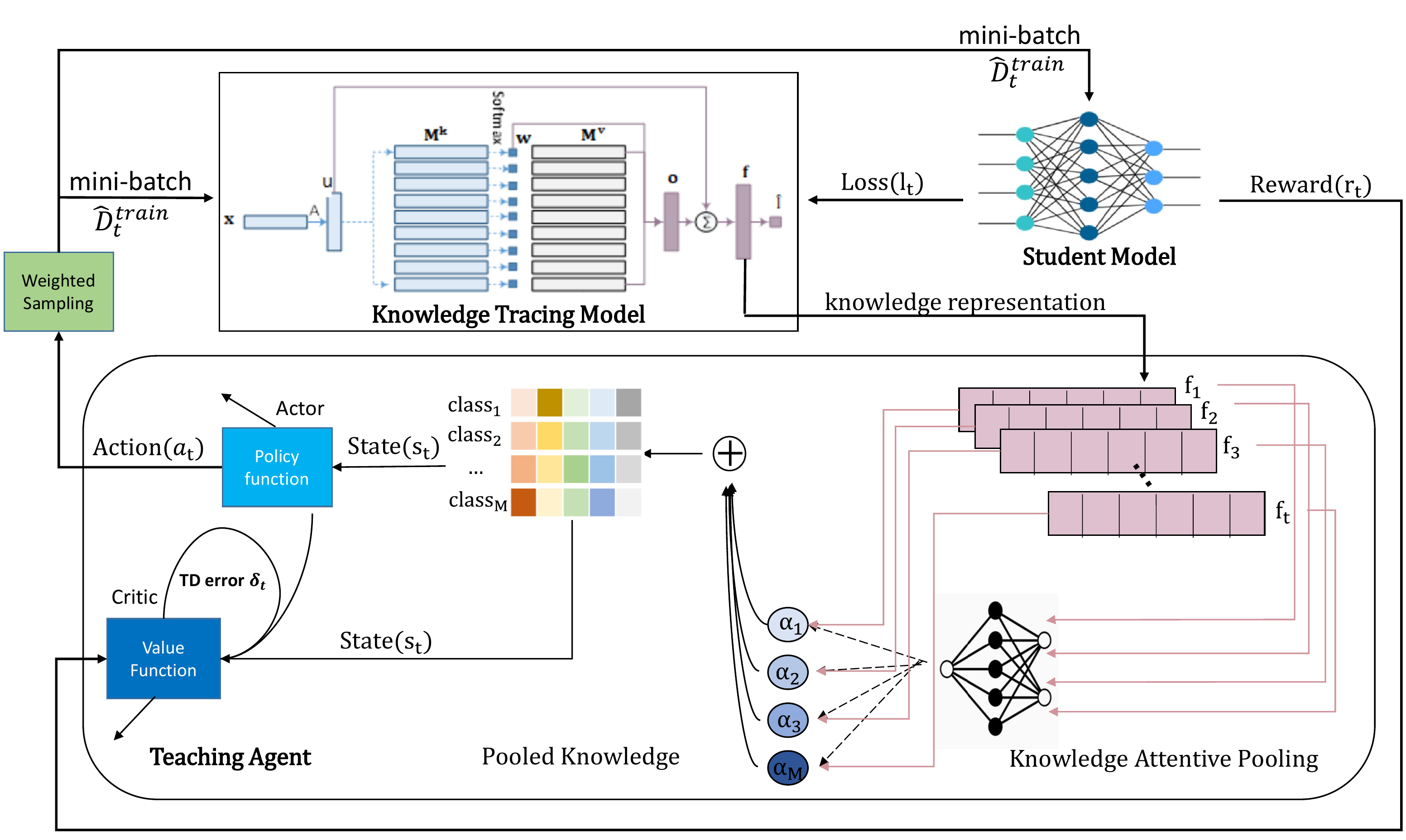}\vspace{-0.1cm}
\caption{Architecture of the proposed KADT method, which involves three main components: a student model, a knowledge tracing model and a teaching agent.}\vspace{-0.3em}
\label{fig:Model}
\end{figure*} 

To address these limitations, our propose a novel framework for developing data teaching strategies, namely \emph{Knowledge Augmented Data Teaching} (KADT). At its core, the KADT method is equipped with a powerful representation learning ability for capturing a student model's performance by leveraging knowledge tracing techniques~\cite{DKT2015,DKVMN17,AbdelrahmanW19}. Specifically, the KADT method employs a key-value memory architecture to learn the knowledge progress of a student model in terms of underlying learning concepts involved in a learning task. This offers several learning advantages: (a) It provides the ability to automatically learn latent learning concepts from training samples in different learning tasks, without explicitly needing any a priori knowledge. (b) It can dynamically track how a student model performs over the learning concepts of a learning task over time (i.e., during the teaching process). 

In addition to these, the KADT method incorporates several novel RL designs, in order to exploit the capacity of a student model and develop a data teaching strategy that matches its capacity to help a student model perform as best as possible. (1) It uses an attentive pooling technique to distill knowledge representations of a student model with respect to class labels through tracing the representations of samples over latent learning concepts. (2) It selects actions directly based on a data teaching strategy, in contrast to comparative RL approaches~\cite{L2T}, where actions depend on the outcome of a preemptive random sampling used to control the complexity of the action space. (3) It uses a dense reward function, which does not require any manual efforts for deciding rewards. However, a sparse reward function as in \cite{L2T} requires choosing a threshold that allows only positive rewards. Manually specifying a good reward threshold is difficult, especially for intricate learning tasks.

\vspace{0.1cm}
\noindent\textbf{Contributions.~}To summarize, the main contributions of this work are below:
\begin{itemize}
    \item We propose a novel teaching method, called KADT, which enables a coupled optimization of knowledge representation learning and teaching strategy learning through the interaction among a student model, a knowledge tracing model and a teaching agent.  
       \item We devise a knowledge representation learning technique that can dynamically trace performance of a student model over different learning concepts in \emph{any} supervised learning task.
       \item We propose an efficient gated attentive pooling mechanism that distills a state representation from pooled knowledge representations of a student model with respect to class labels, while accounting for importance of individual training samples.
 \item We evaluate the KADT method on four different kinds of learning tasks against the state-of-the-art methods. The results empirically validate that the KADT method consistently outperforms the other methods on all the tasks. 
\end{itemize} 

\noindent\textbf{Outline.~}The reminder of this paper is organized as follows.  Section \ref{sec:pd} presents the problem definition. Section \ref{sec:method} describes our methodology. Section \ref{sec:exp} introduces the experimental design. Section \ref{sec:results} discusses the evaluation results. Section \ref{sec:rl} reviews the related work and we conclude the paper in Section \ref{sec:conclusion}.

%% file: Problem.tex
\section{Problem Definition}
\label{sec:pd}
In supervised learning, a learning problem is typically formulated as follows. Given a training set $D^{\mathrm{train}}=\{(\mathbf{x}_i,y_i)\}_{i=1}^{|D^{\mathrm{train}}|}$ consisting of samples $\mathbf{x}_i$ from a sample space $X$ and their labels $y_i$ from a label space $Y$, a hypothesis function $h_{\theta}:$ $X\rightarrow Y$, parameterized by $\theta$, which maps a sample space $X$ to a label space $Y$, and a function $\eta$ that measures the performance of $h_{\theta}$ on $D^{\mathrm{test}}$ after it is trained on $D^{\mathrm{train}}$, a supervised learning problem is to find the optimal parameters of the hypothesis function which maximize the performance:
\begin{equation}
    \theta^*=\argmax_{\theta\,\in\,\Theta}\,\eta(h_{\theta}, D^{\mathrm{train}}, D^{\mathrm{test}})
    \label{eq:student_obj}
\end{equation}
\noindent where $\Theta$ is a parameter search space for the hypothesis function. We call such a hypothesis function \emph{a student model} which can be any supervised machine learning model (e.g., a deep neural network or a simple linear regression model), and Equation \ref{eq:student_obj} represents its optimization objective.

To teach a student model in solving a supervised learning problem (i.e., guide the search for $\theta^*$), an effective data teaching strategy needs to be found. Such a data teaching strategy can be configured through  a sequence of training mini-batches $\mathcal{{D}}=\{\hat{D}^1,\hat{D}^2,\dots,\hat{D}^{|\mathcal{{D}}|}\}$ being sampled from the training set $D^{\mathrm{train}}$. Hence, given a teacher model $g_{\omega}$, parameterized by $\omega$, the teacher model learns to find a good data teaching strategy $g_{\omega}(D^{\mathrm{train}})=\mathcal{D}$ 
for a student model $h_{\theta}$ to maximize its performance, formulated as in the following optimization problem:
\begin{equation}
    \omega^*=\argmax_{\omega\,\in\,\Omega}{\eta(h_{\theta}, g_{\omega}(D^{\mathrm{train}}), D^{\mathrm{test}})}
    \label{eq:opt_curriculum}
\end{equation}
\noindent where $\Omega$ is a parameter search space for the teacher model. 

In this paper, we aim to develop a reinforcement learning framework in which a teacher model can optimise a data teaching strategy dynamically according to the performance of a student model. 


%% file: Methodology.tex
\section{Methodology}\label{sec:overview}
\label{sec:method}
In this section, we present our proposed method, namely \emph{Knowledge Augmented Data Teaching} (KADT). Figure~\ref{fig:Model} illustrates the architecture of KADT. Given a student model that tackles a supervised learning task, KADT uses a knowledge tracing model to trace the knowledge of the student model in performing this supervised learning task, and a teaching agent to optimise a data teaching strategy for the student model in order to maximize its performance. 


\subsection{Student Knowledge Tracing} 
\label{sec:srl} 


We notice that the performance of a student model largely depends on how it acquires knowledge from different learning concepts involved in a task, e.g., movie genres in a classification task for rating movies. Thus, inspired by knowledge tracing methods~\cite{DKVMN17,AbdelrahmanW19}, 
we design a memory-augmented knowledge tracing (KT) model to capture knowledge representation of a student model based on its past interaction history. The KT model has a key-value memory structure $\mathbf{M}=\langle \mathbf{M}^{k}, \mathbf{M}^{v}\rangle$, where $\mathbf{M}^{k}\in\mathbb{R}^{N\times d_{k}}$ is a static matrix, called the \emph{key matrix}, and  $\mathbf{M}^{v}\in\mathbb{R}^{N\times d_{v}}$ is a dynamic matrix, called the \emph{value matrix}. Let $C=\{c_1,\dots, c_N\}$ be $N$ learning concepts underlying the knowledge representation of a student model. Then $\mathbf{M}^{k}$ stores encoding keys that represent learning concepts, and $\mathbf{M}^{v}$ stores the performance information of a student model for each learning concept, which is dynamically changing over time. Thus, the dimension $N$ reflects the number of learning concepts in the task, while $d_k$ and $d_v$ are dimensions of memory slots. We set the slot dimensions $d_{k}=d_{v}=50$ based on our empirical analysis, while the number of slots $N$ is set differently for each dataset (see further discussion in Section \ref{sec:datasets}).

Given a sample ${\mathbf{x}}$ from the training mini-batch $\hat{D}_t^{\mathrm{train}}$ at the time step $t$, we get its embedding vector $\mathbf{u}\in\mathbb{R}^{d_k}$ by embedding its one-hot encoding vector $\mathbf{\delta}(\mathbf{x})\in\mathbb{R}^{|D^{\mathrm{ train}}|}$ with an embedding matrix $\mathbf{A}\in\mathbb{R}^{|D^{\mathrm{ train}}|\times{d_k}}$. Then, a dot product between $\mathbf{u}$ and each key slot $\mathbf{M}^{k}(i)$ in the key matrix $\mathbf{M}^{k}$ is supplied to a $\mathrm{Softmax}$ layer to get the \emph{relevancy vector} $\mathbf{w}\in\mathbb{R}^{N}$: 
\begin{equation}
\label{eq:weightVec}
\mathrm{w}(i)=\mathrm{Softmax}(\mathbf{u}^{\intercal}\mathbf{M}^{k}(i))
\end{equation}
\noindent where $\mathrm{Softmax}(z_{i})=e^{z_{i}}/\sum_{j}e^{z_{j}}$. Intuitively, the relevancy vector $\mathbf{w}$ reflects the relevance between the sample ${x}$ and the learning concepts in $\mathbf{M}^k$.

After calculating the relevancy vector $\mathbf{w}$, the KT model proceeds in two stages. First, it reads from the value matrix $\mathbf{M}^v$ using $\mathbf{w}$ to predict the expected loss of the student model. Second, it updates $\mathbf{M}^{v}$ after acquiring the actual loss from the student model. We call these stages the \emph{read stage} and the \emph{write stage}, respectively, and discuss them further in detail.

\medskip
\noindent\textbf{1) Read Stage:~}In the read stage, the KT model retrieves the student's performance information with regard to the sample $x$ to predict its expected loss. First, the relevancy vector $\mathbf{w}$ is used to calculate a weighted sum from the memory slots of the value matrix $\mathbf{M}^{v}$, which yields the \emph{read vector} $\mathbf{r}\in\mathbb{R}^{d_v}$:
\begin{equation}
\label{eq:readVec}
\mathbf{r}=\sum_{i=1}^{N}\mathrm{w}(i)\mathbf{M}^{v}(i)
\end{equation}

Then, the read vector $\mathbf{r}$ is
concatenated with the embedding vector $\mathbf{u}$ of the sample $x$, 
and then fed to a $\mathrm{Tanh}$ layer to calculate the \emph{representation vector} $\mathbf{f}\in\mathbb{R}^{N}$: 
\begin{equation}
\label{eq:summaryVec}
\mathbf{f}=\mathrm{Tanh}(\mathbf{W}_{1}^{\intercal}\left[\mathbf{r,u}\right]+\mathbf{b_{1}})
\end{equation}where $\mathrm{Tanh}(z_{i})=(e^{z_{i}}-e^{-z_{i}})/(e^{z_{i}}+e^{-z_{i}})$, $\mathbf{W}_1\in\mathbb{R}^{(d_v+d_k)\times{N}}$ is the weight matrix of the $\mathrm{Tanh}$ layer, and $\mathbf{b}_1$ is a bias vector. This representation vector captures both the current knowledge progress of the student model over sample $x$ and the sample information itself.

Finally, we input the representation vector $\mathbf{f}$ to a linear function and take the dot product of the output vector with $\mathbf{\delta}(\mathbf{x})$ to predict the estimated loss $\mathrm{\hat{l}^{student}}$ of the student model for the sample $\mathbf{x}$ as follows:
 \begin{equation}
\label{eq:Final_Prob}
\mathrm{\hat{l}^{student}}=(\mathbf{W}_2^{\intercal}\mathbf{f}+\mathbf{b}_2)\cdot{\mathbf{\delta}(\mathbf{x})}
\end{equation}
\noindent where $\mathbf{W}_2\in\mathbb{R}^{{N}\times|D^{train}|}$ is a weight matrix, $\mathbf{b}_2$ is a bias vector, and $\mathrm{\hat{l}^{student}}$ is a scalar value. 

\medskip
\noindent\textbf{2) Write Stage:~}After the student model predicts the class label $\hat{y}$ of ${\mathbf{x}}$, we acquire the actual loss $\mathrm{l^{student}}=\ell(\hat{y},y)$, where $\ell(\cdot)$ is the loss function of the student model. Then, the value memory $\mathbf{M}^{v}$ is updated to reflect the student model's current performance via the write stage.


Specifically, for each sample $\mathbf{x}$ and its prediction error $\epsilon=|\hat{y}-y|$, we embed them using an embedding matrix $\mathbf{B}\in\mathbb{R}^{(2|D^{\mathrm{ train}}|\times{d_v})}$ to get an embedding vector $\mathbf{j}\in\mathbb{R}^{d_v}$. Afterwards, we concatenate $\mathbf{j}$ with the representation vector $\mathbf{f}$ to add the corresponding performance information. The resulting write vector $\mathbf{v}=[\mathbf{f},\mathbf{j}]\in\mathbb{R}^{N+d_v}$ is used to update the value memory $\mathbf{M}^{v}$ through erase and add signals~\cite{NTM2014}.

Based on the write vector $\mathbf{v}$, we update the value matrix $\mathbf{M}^{v}$. This is done through erasing the existing information from $\mathbf{M}^{v}$ using an erase signal $\mathbf{e}\in\mathbb{R}^{d_v}$, and then adding new information to $\mathbf{M}^{v}$ using an addition signal $\mathbf{a}\in\mathbb{R}^{d_v}$. 

Let $\mathbf{W}_{e}\,\in\mathbb{R}^{(N+d_v)\times d_{v}}$ be a weight matrix and $\mathbf{b}_e$ is a bias vector. The \emph{erase signal} is calculated as follows:
\begin{equation}
 \label{eq:erase_signal}
\mathbf{e}=\mathrm{Sigmoid}(\mathbf{W}_{{e}}^{\intercal}\cdot{\mathbf{v}}+\mathbf{b}_{e})
\end{equation}
\noindent where $\mathrm{Sigmoid}(z_{i})={1}/(1+e^{-z_{i}})$. Let $\mathbf{1}$ be a row vector of all ones, the updated value matrix $\tilde{\mathbf{M}}_{\mathrm{updated}}^{v}$ is calculated using element-wise multiplication as follows:
\begin{equation}
\label{eq:Erased_Memory}
\tilde{\mathbf{M}}_{\mathrm{updated}}^{v}(i)=\mathbf{M}^{v}(i)[\mathbf{1}-\mathrm{w}(i)\mathbf{e}]
\end{equation}

\noindent Thus, the $i_{\mathrm{th}}$ slot of $\mathbf{M}^{v}$ is erased (i.e., set to zero) if the corresponding values in the relevancy vector $\mathbf{w}$ and the erase signal $\mathbf{e}$ are both equal to one, and remains unchanged if either of them is zero.

Let $\mathbf{W}_a\,\in\mathbb{R}^{(N+d_v)\times d_{v}}$ be a weight matrix. The \emph{addition signal} $\mathbf{a}$ is calculated as below:
\begin{equation}
\label{eq:Add_signal}
\mathbf{a}=\mathrm{Tanh}(\mathbf{W}_{a}^{\intercal}\cdot\mathbf{v+b_{a}})
\end{equation}  

 Finally, the value matrix $\mathbf{M}^{v}$ is updated as:
 \begin{equation}
\mathbf{M}^{v}(i)=\tilde{\mathbf{M}}_{\mathrm{updated}}^{v}(i)+\mathrm{w}(i)\mathbf{a}
\end{equation}

\medskip
\noindent\textbf{3) Model Optimization:~}To optimize the KT model's parameters (i.e., the embedding matrices, weight and bias parameters of different neural layers), we utilize the \emph{Root Mean Square Error} (RMSE) function to calculate the differences between the estimated loss  $\mathrm{\hat{l}^{student}}$ and the actual loss $\mathrm{l^{student}}$, acquired after the student model predicates the class label for each sample in $\hat{D}_t^{\mathrm{train}}$. 

The KT model is trained using the following loss function: 

\begin{equation}
\label{eq:srl-loss}
\mathcal{L_{\mathrm{KT}}\mathrm{=\sqrt{\frac{\sum_{i=1}^{|\hat{D}^{\mathrm{train}}|}\left (\hat{l}^{student}_{i}-l^{student}_{i}  \right )^{2}}{|\hat{D}^{\mathrm{train}}|}}}}
\end{equation}

After calculating the RMSE error based on the current training mini-batch, the parameters of the KT model are updated using gradient decent through back-propagation.


\subsection{Data Teaching Strategies}\label{sec:teacher}
In this work, we design a teacher model in a reinforcement learning framework, called \emph{teaching agent}. The teaching agent aims to optimize a \emph{teaching policy} (i.e., a data teaching strategy) for a student model guided by the knowledge representation of a student model learnt by the KT model. 


\medskip
\noindent\textbf{1) Teaching Interactions:~}
Following the reinforcement learning paradigm~\cite{MDP,bellman1957dynamic}, we model teaching interactions as a Markov decision process (MDP), represented by a tuple $(S,A,T,R)$. Here, $S$ is a state space, $A$ is an action space, $T: S\times A \times S \rightarrow [0, 1]$ is a state transition function such that $T(s_t,a_t,s_{t+1})=P(s_{t+1}|a_t,s_t)$ represents transition probabilities between states after executing actions, and $R: S\times A\rightarrow \mathbb{R}$ is a reward function. Below, we discuss them in detail.


\vspace*{0.2cm}
\noindent\emph{States.~}Each state $s_t\in S$ in our work is represented as a matrix $s_t\in\mathbb{R}^{O\times{N}}$, where $O$ is the number of class labels and $N$ is the number of learning concepts in a learning task. Intuitively, each row in this matrix represents the knowledge of a student model corresponding to a class label, which is pooled from the representation vectors of its training samples w.r.t. the underlying learning concepts (i.e., columns).

Specifically, we distill the latest knowledge representation of a given class label $y$ by calculating a \emph{pooled knowledge vector} $g^{y}_{t}\in \mathbb{R}^N$ from the knowledge vectors of its training samples in the lately sampled mini-batch $\hat{D}^{train}_{t-1}$. The knowledge vectors of a mini-batch are projected by the read stage of the KT model (see Section~\ref{sec:srl}). In the first teaching interaction, the mini-batch $\hat{D}^{train}_{t-1}$ is sampled using a uniform random distribution across all class labels and training samples (i.e., sampling from all class labels with an equal probability and treating samples in each class label as being equally likely); afterwards, it is the lately sampled mini-batch by our teaching agent. We follow an attentive pooling method to calculate $g^{y}_{t}$ as follows:

\begin{equation}
    \label{eq:pooledKS}
    g^{y}_{t} = \sum_{i=1}^{I} \alpha^{i}_{t}\mathbf{f}^{i}_{t}
\end{equation}

\noindent where $I$ is the total number of training samples belonging to class label $y$ in the mini-batch $\hat{D}^{train}_{t-1}$, $\alpha^{i}_{t}\in[0,1]$ is the attention weight for sample $\mathbf{x}_i$ at time point $t$, and $\mathbf{f}^{i}_{t}$ is the knowledge representation vector of sample $\mathbf{x}_i$ at time point $t$ calculated as per Equation~\ref{eq:summaryVec}.

The attention weight $\alpha^{i}_{t}$ for sample $\mathbf{x}_i$ is calculated using a gated attention mechanism~\cite{Bahdanau14, IlseTW18}, which controls the amount of information to pass from each sample to the pooled knowledge vector of its corresponding class label:

\begin{equation}
    GA(\mathbf{f}^{i}_{t})=\exp(\mathbf{c}^{\intercal}\frac{(Tanh(\mathbf{W}^{\intercal}\mathbf{f}^{i}_{t})\odot\,Sigmoid(\mathbf{U}^{\intercal}\mathbf{f}^{i}_{t})}{\sqrt{N}})
\end{equation}
\begin{equation}
    \label{eq:pooling_attention}
    \alpha^{i}_{t}=\frac{GA(\mathbf{f}^{i}_{t})}{\sum_{j=1}^{J}{GA(\mathbf{f}^{j}_{t})}}
\end{equation}

\noindent where $\mathbf{c}\in\mathbb{R}^{L}$ is a learnable weight vector, $\mathbf{W}\in\mathbb{R}^{N\times{L}}$ is a learnable weight matrix for the $Tanh$ function, $\mathbf{U}\in\mathbb{R}^{N\times{L}}$ is a learnable weight matrix for the $Sigmoid$ function, operator $\odot$ is an element-wise product, $\sqrt{N}$ is a scaling factor to control the output range and prevent gradient vanishing~\cite{Vaswani_17}, and $J$ is the total number of training samples belonging to the class label in the mini-batch. The $Sigmoid$ function acts as a gating filter to control the information to pass from the $Tanh$ function.

\vspace*{0.2cm}
\noindent\emph{Actions.~}Given a state, the teaching agent selects a mini-batch of training samples and sends this mini-batch to the student model to train on during each interaction. An action $a_t\in \mathbb{R}^{O}$ is a \emph{sampling vector} normalized using a $Softmax$ output layer, satisfying:

\begin{equation}
\label{eq:readVec}
\sum_{i=1}^{O}{a}_t(i)=1.
\end{equation}
Each $a(i)$ corresponds to a class label, and the value of $a(i)$ indicates the percentage to sample from the corresponding class label in the next mini-batch. To sample the next mini-batch, we utilize a weighted sampling method that takes the action $a_t$ and the latest attention weights over training samples in each class label and outputs the mini-batch $\hat{D}^{train}_{t}$. Considering attention weights during sampling counts for important samples that could advance the learning progress of a student over the coming teaching interactions.

As we follow an efficient attentive pooling method that distills the knowledge representation of a given class label only using its training samples in the lately sampled mini-batch, we depend on an estimation technique to get the attention weights of its remaining samples that were not lately sampled. Our technique estimates the weights of other samples in class label $y$ through a moving average given the weights of the present samples as follows:

\begin{equation}
    \label{eq:mov_avg_weights_part1}
    \hat{\alpha}^{u}=\sum_{i=1}^{I} (x^u\cdot{x^i})\times{\alpha^{i}_{t}}\,\,\text{ for }x^u\not\in\hat{D}^{train}_{t-1}
\end{equation}

\begin{equation}
    \label{eq:mov_avg_weights_part2}
    \alpha^{u}_{t}=\frac{\alpha^{u}_{t-1}+\hat{\alpha}^{u}}{\Gamma}
\end{equation}

\noindent where $\hat{\alpha}^{u}$ is a weighted sum estimate of attention weight for sample $x^u$ in class label $y$, $I$ is the total number of training samples belonging to class label $y$ in the mini-batch $\hat{D}^{train}_{t-1}$, and $\Gamma$ is the number of previous teaching interactions.

\vspace*{0.2cm}
\noindent\emph{Reward Function:~}
Let $p_t=\eta(h_{\theta^*}, \hat{D}^{train}_{t}, D^{\mathrm{valid}})$ denotes the performance of a student model $h_{\theta^*}$ at the time step $t$ after being trained on the mini-batch $\hat{D}^{train}_{t}$, and validated on validation set $D^{\mathrm{valid}}$.
Then, our reward function $R$ is calculated to reflect the magnitude of the performance change for a student model defined as the below piecewise function :
  \begin{equation}
 \label{eq:reward_function}
    R(s_t,a_t) =\begin{cases} 
0 & \text{if }|p_{t}-p_{t-1}|\leq\epsilon; \\
       (p_{t}-p_{t-1}) &\text{otherwise}.
   \end{cases}
 \end{equation}
In order to prevent oscillations over small variations, this reward function is designed to be $\epsilon$-insensitive to the magnitude of the performance change, where $\epsilon \leq 0.1$ is a hyper-parameter.

The key idea behind this reward function is to positively reward actions that can improve the learning progress of a student model and penalize those that slow down the progress. The effectiveness of using the learning progress as a reward signal has been previously studied in 
the RL literature~\cite{Oudeyer_2007, Schmidhuber_2010}. In our work, enhancing the learning progress needs \emph{balanced mini-batches} that contain new or challenging samples in addition to known ones to prevent forgetting. For example, selecting training mini-batches that have been correctly classified by a student model will not enhance the performance; hence, our reward function will generate a low reward value, i.e., zero, since it will lead to the same performance. On the other hand, selecting only hard training mini-batches will also decrease the performance of a student model in the long run and our reward function yields low reward values accordingly. By setting the reward function around the learning progress, we do not need to use heuristic rules (e.g., hardness thresholds) that require domain knowledge and might bias the learning process toward specific samples.

\vspace{0.2cm}
\noindent\emph{Remark.~}
There are two possible ways for designing our reward function: sparse reward function and dense reward function. A sparse function, as used in L2T~\cite{L2T}, requires to specify a performance threshold. Thus, only values that exceed the performance threshold yield positive reward signals; otherwise, the reward function would give zero. However, setting a performance threshold is a difficult task in real-world applications, and an unrealistic performance threshold would negatively affect the effectiveness of a teaching policy. Moreover, learning with sparse reward signals often leads to unstable behaviors and slower convergence, in comparison with using a dense reward function~\cite{rl2018}. Therefore, we design a dense reward function that maps the performance change of a student model into a reward value, rather than using a sparse reward function with a pre-defined performance threshold.

\medskip
\noindent\textbf{2) Actor-Critic Algorithm:~}We design our RL algorithm based on deep deterministic policy gradient (DDPG)~\cite{DDPG2016}, consisting of two building blocks: (1) \emph{a critic network} $Q(s,a\,|\,\theta^{Q})$ estimates the Q-value of the current state-action pairs; (2) \emph{an actor network} $\psi(s\,|\,\theta^{\psi})$ that learns to select an optimal action for the current state.

Following DDPG, we use the technique of reply buffer (RB) to build a mini-batch $V$ of (state, action, reward) transitions which updates the parameters of the critic and actor networks through gradient decent. We also follow the concept of target networks which are versions of the critic and actor networks with parameter values $(\theta^{Q^{\prime}}, \theta^{\psi^{\prime}})$ being updated proportionally to the latest actor and critic values $(\theta^{Q}, \theta^{\psi})$ with a delay factor $\tau<\!< 1$ as shown in Equations~\ref{eq:target_critic} and \ref{eq:target_actor}:
\begin{equation}
\label{eq:target_critic}
    \theta^{Q^{\prime}\leftarrow}  \tau\theta^{Q}+(1-\tau)\theta^{Q^{\prime}}
\end{equation}\vspace*{-0.3cm}
\begin{equation}
\label{eq:target_actor}
    \theta^{\psi^{\prime}}\leftarrow  \tau\theta^{\psi}+(1-\tau)\theta^{\psi^{\prime}}
\end{equation}
The purpose of these target networks is to make the optimization of the actor and critic networks stable as we calculate the gradient updates based on the Q-value estimate from the critic network itself~\cite{DDPG2016}.

The critic network's parameters are optimized to minimize the temporal difference (TD) loss~\cite{rl2018} shown in Equation~\ref{eq:TD_Error}. The actor network's parameters along with the attention learnable parameters are updated through the policy gradient function~\cite{DDPG2016} depicted in Equation~\ref{eq:pg}.
\begin{multline}
L_{TD}=\\\frac{1}{|V|}\,\sum_{i=1}^{|V|}(Q(s_{i},a_{i}|\theta^{Q}) - (r_{t+1}+\gamma\max_{a^{'}}Q\,(s_{t+1},a^{\prime}|\theta^{Q^{\prime}}))^{2}
\label{eq:TD_Error}
\end{multline}
\begin{equation}
 \nabla_{\theta^{\psi}}J \thickapprox\\\frac{1}{|V|}\,\sum_{i=1}^{|V|}\nabla_{a}Q(s,a\,|\theta^{Q})\,|_{s=s_{i},a=\psi(s_{i})}\nabla_{\theta^{\psi}}\psi(s\,|\theta^{\psi})\,|_{s_{i}} 
 \label{eq:pg}
\end{equation}
\noindent where $\gamma$ is the discounting factor and $|V|$ is the length of a transitions mini-batch sampled from the reply buffer.

The action exploration is performed randomly using the \textit{Ornstein–Uhlenbeck} (OU) stochastic process~\cite{DDPG2016}, which generates temporally correlated exploration noise for a smooth transition across action values. The OU process is calculated as per Equation \ref{eq:OU}: 
\begin{equation}
da_{t}=\theta(\beta-a_{t})dt+\sigma dW_{t}
\label{eq:OU}
\end{equation}
\noindent where $\theta$, $\sigma$, and $\beta$ are parameters. $W_{t}$ represents the Wiener process~\cite{DDPG2016}, which is a stochastic process being initialized as $W_{0}=0$ and incremented by a Gaussian random value $(W_{t_{i}}-W_{t_{j}})\sim\mathcal{N}(0,t_{i}-t_{j})\,\forall\, 0\leq t_{j}<t_{i}$ at each time step. Algorithm \ref{alg:CAL2T} describes the main steps for training our KADT method.

\begin{algorithm}
\caption{Training KADT} \label{alg:CAL2T}
\begin{algorithmic}[1]
\State{Initialize critic ${Q(s,a\,|\,\theta^{Q})}$ with ${\theta^{Q}}$ at random}
\State{Initialize actor ${\psi(s\,|\,\theta^{\psi})}$ with ${\theta^{\psi}}$ at random}
\State{Initialize target networks ${Q^{\prime}}$ and ${\psi^{\prime}}$ with  ${\theta^{Q^{\prime}}\leftarrow\theta^{Q}}$ and ${\theta^{\psi^{\prime}}\leftarrow\theta^{\psi}}$}
\State{Initialize KT parameters at random}
\State{Initialize reply buffer RB}
\For{episode = 1, ${M}$}
        \State{Initialize OU process ${da_{t}}$ for action exploration}
        \State{Let $F_t\leftarrow{KT.read(\hat{D}_{t-1}^{train})}$}
        \State{Let $s_t\leftarrow{concat(g^{y}_{t})}\forall{y\in{Y}}$ }
        \For{${t = 1}$, ${T}$}
            \State{Select ${a_t=\psi(s_t\,|\,\theta^{\psi})+da_{t}}$}
            \State{Estimate $\mathbf{\alpha}_t$ using Equation~\ref{eq:mov_avg_weights_part2}}
            \State{${\hat{D}_{t}^{train}}\leftarrow{WeightedSample}({a_t,\mathbf{\alpha}_t,D^{train}})$} 
            \State{Train the student on ${\hat{D}_{t}^{train}}$}
            \State{Observe reward ${r_t}$ and new state ${s_{t+1}}$}
            \State{Update the student's parameters using SGD}
            \State{Update the KT parameters using Equation~\ref{eq:srl-loss}}
            \State{Save transition ${(s_t,a_t,r_{t},s_{t+1})}$ into RB}
            \State{Sample a random batch of $V$ transitions from RB} 
            \State{Update critic using ${L_{TD}}$ in Equation~\ref{eq:TD_Error}}
            \State{Update actor using policy gradient in Equation~\ref{eq:pg}}
            \State Update target networks using Equations~\ref{eq:target_critic}-\ref{eq:target_actor}
        \EndFor
        \State \textbf{endfor}
\EndFor
\State \textbf{endfor}
\end{algorithmic}
\end{algorithm}

\medskip
\noindent\textbf{3) Model Optimization:~}
The optimization objective of the teaching agent is to find an optimal teaching policy $\pi^*:S \rightarrow A$ that maximizes the average reward gain: 
 \begin{equation}
 \label{eq:max}
 \pi^*=\argmax_{\pi\,\subset\,\Pi}\,\frac{1}{E}\sum_{i=1}^{E}{r^{\pi}_{i}}
\end{equation}   

\noindent where $\Pi$ is a teaching policy search space and $E$ is the total number of teaching episodes. We optimize the actor-critic parameters $(\theta^{Q},\theta^{\psi})$ using Equations~\ref{eq:TD_Error}
and \ref{eq:pg} to find the optimal teaching policy $\pi^{*}$.

%% file: Experiments.tex
\begin{table*}[ht]
  \caption{Dataset statistics and the corresponding student model setup, where the sizes of mini-batches are set based on empirical analysis on validation sets.}\vspace{-0cm}
  \label{tbl:datasets}
  \begin{center}
  \resizebox{\textwidth}{!}{\begin{tabular}{c|c|c|c|c|c|c}
  \hline
    \toprule 
  \multirow{2}{*}{\textbf { Task}}&\multirow{2}{*}{\textbf{Dataset}}&\multirow{2}{*}{\textbf{Size}}&\multirow{2}{*}{\textbf{\#Classes}} & \multicolumn{2}{|c|}{\textbf{{Student Model Setup}} \hspace{0.5cm}\textbf{[training$\rightarrow$Testing]}}& \textbf{Mini-batch}  \\ \cline{5-6} 
    &&&& \textbf{Similar Student Mode}\hspace{0.3cm} & \textbf{\hspace{0.3cm}Different Student Mode }&\textbf{Size}\\
     \midrule
    Knowledge Tracing&ASSISTments2009&$325,637$&$2$ & [SKVMN$\rightarrow$SKVMN] & [SKVMN$\rightarrow$DKT]&256\\
    \hline
    Sentiment Analysis&IMDB&$50,000$&$2$ &[LSTM$\rightarrow$LSTM]& [LSTM$\rightarrow$SVM] & 16\\\hline
    Movie Recommendations \hspace*{0.2cm}&MovieLens&$1000,000$&$2$ &[LSTM$\rightarrow$LSTM]&[LSTM$\rightarrow$MF] &1000\\\hline
    Image Recognition&CIFAR-100&$60,000$&$100$ &[MLP$\rightarrow$MLP] &[MLP$\rightarrow$CNN]&128\\
  \bottomrule
\end{tabular}}
\end{center}
\end{table*}

\section{Experiments}
\label{sec:exp}
We conduct experiments to evaluate our proposed KADT method against the state-of-the-art methods. These experiments aim to answer the following research questions:
\begin{description}
    \item[RQ1:] How effectively can our KADT method teach a student model across different learning tasks?
    \item[RQ2:] How well can our KADT method be generalized to teach different student models, i.e., the generalization ability of our teaching agent?
    \item[RQ3:] How well can our KADT method learn knowledge representation of a student model to improve the performance?
    \item[RQ4:] How well can our KADT method perform in comparison to the state-of-the-art RL teaching method?
    \item[RQ5:] How does each main component of our KADT method affect its performance?
\end{description}

Below, we introduce our experimental setup for answering the above research questions.

\begin{table*}[htbp]
  \caption{Accuracy results for teaching with similar
   and different student modes, averaged over 5 runs. }\vspace{-0cm}
  \label{tbl:accuracy_modes}
  \begin{adjustbox}{max width=\textwidth}
  \begin{tabular}{c|c|c|c|c|c|c|c|c}
   \hline
    \toprule
   \multirow{2}{*}{\textbf{{Dataset}}}& \multicolumn{4}{c|}{\textbf{Similar Student Mode}}& \multicolumn{4}{c}{\textbf{Different Student Mode}}  \\\cline{2-9}
    &\textbf{RandomTeach}&\textbf{SPL}&\textbf{L2T}&\textbf{{KADT}}&\textbf{RandomTeach}&\textbf{SPL}&\textbf{L2T}&\textbf{{KADT}}\\ 
     \midrule
    ASSISTments2009&$81.63\pm0.4$&$83.45\pm0.02$&$84.31\pm0.03$&$\mathbf{87.10\pm0.04}$&$79.82\pm0.3$&$80.41\pm0.06$&$81.29\pm0.04$&$\mathbf{84.26\pm0.04}$\\\hline
    IMDB&$88.54\pm0.8$&$88.80\pm0.05$&$89.46\pm0.06$&$\mathbf{92.24\pm0.05}$&$78.87\pm0.5$&$81.04\pm0.03$&$83.22\pm0.05$&$\mathbf{86.16\pm0.03}$\\\hline
    MovieLens&$75.12\pm0.6$&$76.45\pm0.06$&$77.23\pm0.03$&$\mathbf{80.31\pm0.02}$&$74.82\pm0.8$&$76.87\pm0.04$&$77.73\pm0.06$&$\mathbf{80.47\pm0.03}$\\\hline
    CIFAR-100&$62.18\pm0.4$&$64.68\pm0.04$&$65.33\pm0.06$&$\mathbf{68.75\pm0.03}$&$68.71\pm0.5$&$70.27\pm0.05$&$71.82\pm0.03$&$\mathbf{74.58\pm0.02}$\\
  \bottomrule
\end{tabular}
\end{adjustbox}
\end{table*}

\subsection{Datasets}
\label{sec:datasets}
We consider four different kinds of learning tasks in our experiments: a knowledge tracing task, a sentiment analysis task, a movie recommendation task, and an image classification task. For these tasks, we use four different datasets as follows (see Table~\ref{tbl:datasets}):  

\smallskip
\noindent\textbf{Knowledge Tracing on ASSISTments2009 \footnote{\href{https://sites.google.com/site/assistmentsdata/home/assistment-2009-2010-data}{ASSISTments2009 Source}}}
The dataset ASSISTments2009 was collected during the school year $2009-2010$ using the ASSISTments online education website \footnote{\url{https://new.assistments.org/}}. It consists of $110$ distinct questions answered by $4,151$ students which leads to a total number of $325,637$ exercises. The questions are represented as one-hot encoding vectors (i.e., one value in the corresponding index of each question in the dataset and zeros elsewhere). Based on previous studies conducted on this dataset~\cite{AbdelrahmanW19,DKVMN17}, there are $10$ different learning concepts in the questions. There are two class labels in this dataset including \emph{correct} and \emph{incorrect}.

\smallskip
\noindent\textbf{Sentiment Analysis on IMDB\footnote{IMDB: \url{http://ai.stanford.edu/~amaas/data/sentiment/}}:}
The IMDB~\cite{Maas11} dataset includes movie reviews in text with a total of $50,000$ reviews. There are two class labels: \emph{positive review} and \emph{negative review}. The overall distribution of labels is balanced, with $25k$ for the positive review and $25k$ for the negative review. The text reviews are embedded using the word2vec embedding with dimension size of 256. There are $11$ learning concepts representing the general categories of movies reviewed in the text.

\smallskip
\noindent\textbf{Movie Recommendations on MovieLens \footnote{MovieLens: \url{https://grouplens.org/datasets/movielens/}}:}
The MovieLens dataset includes $1$ million movie ratings performed by $6000$ users on $4000$ movies, released in $2003$. Each move rating has three features: user id, movie id, and timestamp. There are five class labels representing a five star rating (i.e., from 1 to 5), and $18$ learning concepts representing the movie genres. We follow the setting of Click-Through Rate (CTR) (i.e., binary labels: click and no click) by regarding labels 4 and 5 as click and labels 1 to 3 as no-click. This allows us to standardize this evaluation into a classification task.

\smallskip
\noindent\textbf{CIFAR-100 Image Classification \footnote{CIFAR-100:\url{https://www.cs.toronto.edu/~kriz/cifar.html}}:} The CIFAR-100 dataset was collected by ~\cite{CIFAR-100}. The dataset consists of $60000$ RGB images of $32x32$ pixels. There are $100$ class labels with $600$ images for each class. These $100$ classes in CIFAR-100 are grouped into $20$ super-classes, and each super-class corresponds to a learning concept. Thus, we use $20$ learning concepts in this dataset.

\subsection{Baseline Methods}\label{subsec:baselines}

We discuss the baseline methods used in the teacher model setup and the student model setup in our experiments.

\medskip

\noindent\textbf{Teacher Model Setup.~}We compare the performance of the proposed method KADT against two state-of-the-art methods: 
\begin{itemize}
\item\textbf{Self-Paced Learning (SPL)~\cite{KumarPK10}:} This method works by filtering out training examples that have a loss value exceeding a predefined hardness threshold. This threshold value increases gradually during the training time.
\item\textbf{Learning to Teach (L2T)~\cite{L2T}:} This method uses a reinforcement learning agent to sample training mini-batches for a student model. The agent optimizes a sparse reward function that yields a positive reward only when the student model's performance exceeds a predefined performance threshold.
\end{itemize}
In addition, we use random sampling as a baseline model and refer to it as ``RandomTeach''
\begin{itemize}
\item\textbf{Random Sampling (RandomTeach):} This method provides the conventional data teaching strategy, i.e., sampling mini-batches from a uniform random distribution. 
\end{itemize}

For the SPL and L2T methods, we follow the same parameter configuration as described in~\cite{KumarPK10} and~\cite{L2T}, respectively. Each of these methods has an additional hyper-parameter, i.e., the hardness threshold in SPL and the performance threshold in L2T.  For SPL, we use a range of values $\{80,100,120,150,180\}$ for the hardness threshold. For L2T, we follow the heuristic strategy in~\cite{L2T} that sets the best performance value from the past episodes as the threshold for the following ones. We report their best results.

\begin{figure*}
  \includegraphics[width=\textwidth]{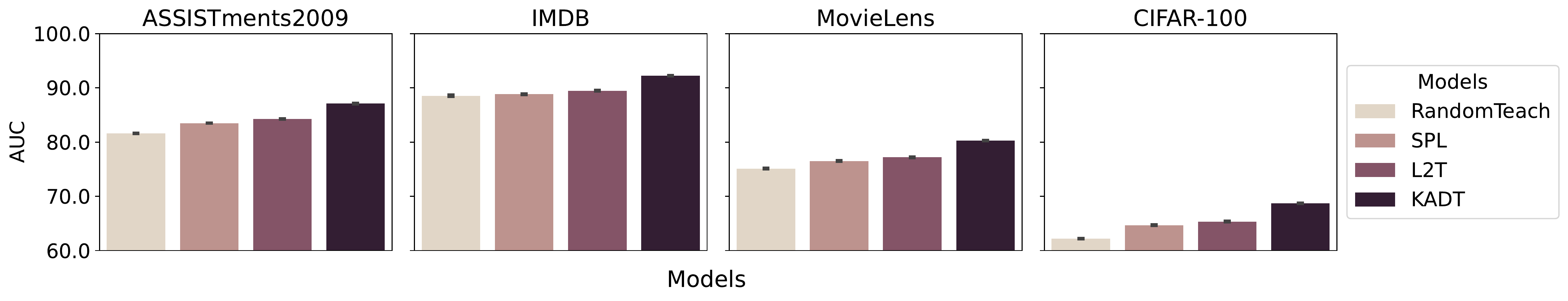}\vspace{-0.2cm}
   \caption{AUC results averaged over 5 runs for the teaching with same student mode over four datasets.}
   \label{fig:roc_similar}
  \vspace{-0.2cm}
\end{figure*}
\begin{figure*}
   \includegraphics[width=\textwidth]{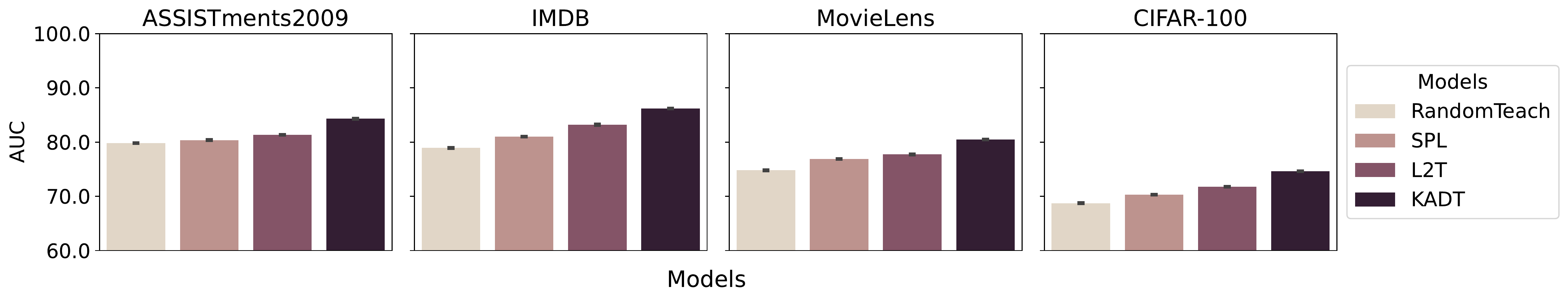}\vspace{-0.2cm}
\caption{AUC results averaged over 5 runs for the teaching with different student mode over four datasets.}
   \label{fig:roc_different}
\vspace{-0cm}

\end{figure*}

\medskip
\noindent\textbf{Student Model Setup.~} We consider two different student models for each dataset. For the knowledge tracing task, we use deep knowledge tracing (DKT)~\cite{DKT2015} and sequential key-value knowledge tracing (SKVMN)~\cite{AbdelrahmanW19}. For the sentiment analysis task, we use a support vector machine with a radial basis kernel (SVM) and a long short-term memory (LSTM)~\cite{LSTM_Student} model. For the recommendation task, we use a matrix factorization (MF) model~\cite{MF8} and the same LSTM model from the sentiment analysis task. For the image classification task, we use a multi-layer perceptron (MLP) and a convolutional neural network (CNN)~\cite{CNN12}. We follow the same design and hyper-parameter configuration as per the cited work for these models.


\subsection{Training and Testing Strategies}
We divide each dataset into a train-validate set and a test set using a $(70\%-30\%)$ ratio. Then, the train-validate set is further divided into $\mathrm{D^{train}}$ and $\mathrm{D^{valid}}$ using a 5-fold cross validation. 

\medskip
\noindent\textbf{Training process.~}The training process consists of two phases in our experiments.

\begin{itemize}
    \item \emph{Phase 1 -- Teacher Training.~} The purpose of this phase is to train a teacher model for learning a data teaching strategy. This is achieved by assigning a fixed number of \emph{350} training episodes, which is selected through empirical analysis. Each training episode has {50} time steps (i.e., episode time horizon) for the L2T and KADT methods and an equivalent number of training epochs for the SPL model. For example, in the IMDB dataset, we have a total of {50,000} samples and then the size of the train-validate set is {35,000}. Following 5-fold cross validation, the number of training samples is {28,000}, which leads to a mini-batch size of {16}. The equivalent number of epochs for {350} episodes is {10} epochs. Note that RandomTeach does not have this phase because there are no parameters to optimize for its data teaching strategy. For the reward calculation in the L2T and KADT methods, we use 5\% of the validation data.

\item\emph{Phase 2 -- Student Training.~} The purpose of this phase is to apply the data teaching strategy learned in Phase 1 on a new student model. There are two different modes: (a) using a student model that is the same as the student model in Phase 1 but with re-initialized parameters (see more details in Section~\ref{sec:mode1}), and (b) using a student model that is different from the student model in Phase 1 (see more details in Section~\ref{sec:mode2}). In this phase, the teacher model is not allowed to update its parameters (i.e., fix the teaching strategy). We give the same number of episodes and epochs as per Phase 1 to apply the data teaching strategies on the student model.
\end{itemize}
\noindent\textbf{Testing process.~}In the testing process, we evaluate the performance of a teacher model by applying the student model trained by its data teaching strategy in Phase 2 on the test set $\mathrm{D^{test}}$. 


\subsection{Student Evaluation Modes}
To answer RQ1 and RQ2, we design our experiments in two modes: \textit{teaching with similar student mode} and \textit{teaching with different student mode}. We present the details of these two modes below:

\vspace*{0.2cm}
\noindent\textbf{Teaching with similar student mode.}
\label{sec:mode1}
In this mode, we use the same type of student models (e.g., SVM) across Phases 1 and 2. To evaluate how effectively a data teaching strategy is learned from the training phase, we randomly initialize the parameters of a student model in the testing phase. This mode evaluates how well the same student model can be taught on new samples (i.e., test samples) from the same dataset.

\vspace*{0.2cm}
\noindent\textbf{Teaching with different student mode.}
\label{sec:mode2}
In this mode, we change the types of student models across Phases 1 and 2. For example, if we train a feed-forward neural network student model, then we test this model by replacing it with a recurrent neural network student model. This mode aims to evaluate the generalization of data teaching strategies optimised by a teacher model over different student models. 

\vspace{0.2cm}


%% file: Results.tex
\begin{figure*}[t!]
\includegraphics[width=\textwidth]{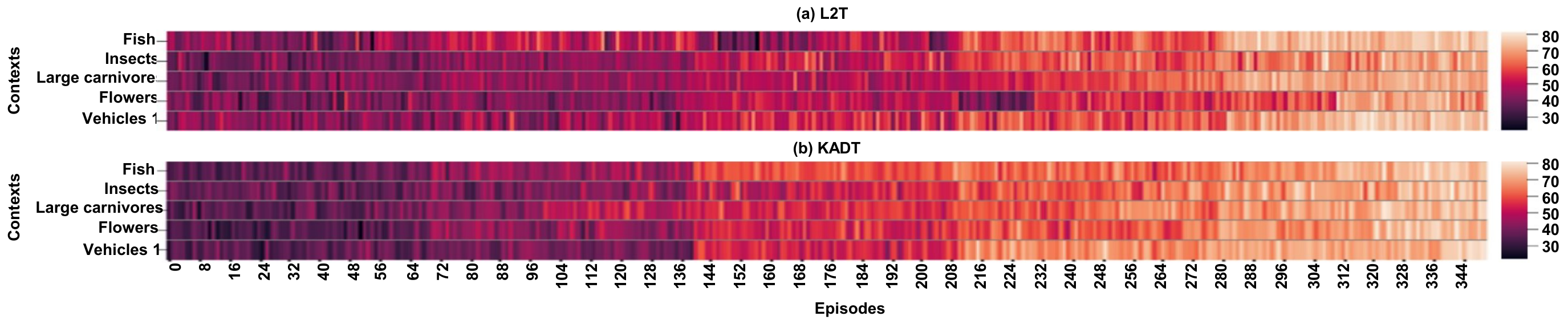}
\vspace{-0.3cm}
\caption{Heatmaps for the validation accuracy for a CNN student model on the CIFAR-100 dataset over five learning concepts. (a) L2T results. (b) KADT results.}
\label{fig:context}
\end{figure*}

\section{RESULTS AND DISCUSSION}
\label{sec:results}

In this section, we present the results of our experiments and discuss our observations. We answer the research questions RQ1 - RQ5 in Section~\ref{sec:teach_similar} - Section~\ref{sec:AS}, respectively. 

\subsection{Teaching with Similar Student Models}
\label{sec:teach_similar}

Table \ref{tbl:accuracy_modes} shows the average classification accuracy achieved by the student models in the teaching with similar student mode. It can be observed that our KADT method outperforms the other methods on all datasets with a statistically significant margin (p-value $<$ $0.05$), and it is followed by the L2T method. This shows the effectiveness of using reinforcement learning to learn a data teaching strategy in comparison to the other methods. Further, the models with learnable teaching strategies (i.e., KADT, L2T and SPL) perform better than RandomTeach that uses the random teaching strategy. The performance of KADT gained above L2T demonstrates the effectiveness of our KT model in learning knowledge representations, in comparison to using hand-crafted states in L2T. Overall, KADT outperforms the second best-performed method L2T by a margin of $2.79$, $2.78$, $3.08$, and $3.42$ on the datasets ASSISTments2009, IMDB, MovieLens, and CIFAR-100, respectively. The performance margin on CIFAR-100 was the smallest across all datasets because CIFAR-100 is the most challenging dataset with $100$ different classes and the other datasets have binary classes.

Figure \ref{fig:roc_similar} shows the AUC results, which confirm the above findings. Our KADT method achieves the highest AUC value across the four supervised learning tasks. 

\subsection{Teaching with Different Student Models}
\label{sec:teach_different}

Table \ref{tbl:accuracy_modes} also shows the average classification accuracy for the student models in the teaching with different student mode. Similarly, our KADT method outperforms the L2T model 
(the next best performer) by a margin of $2.97$, $2.94$, $2.74$, and $2.76$. 
For the ASSISTments2009 and IMDB results, a noticeable performance degradation occurs due to transforming from student models (i.e., SKVMN and LSTM respectively) to less capable models (i.e., DKT and SVM respectively). For the IMDB dataset, the results are comparable to the previous experiment as the learning capacity of the two student models (i.e., LSTM and MF) is similar in the recommendation task. A significant enhancement can be observed in the results of the CIFAR-100 dataset in comparison to the previous experiment. This is because the image representation capacity of the CNN student model used in this experiment is better than the one using the MLP model explored previously. Despite these changes in the performance results, it can still be observed that the learnable teaching methods ((i.e., KADT, L2T and SPL)) enhance the performance in comparison to the RandomTeach method. Figure \ref{fig:roc_different} confirms the performance gain achieved by KADT method that outperforms the next best performer L2T on the AUC metric by margins of $2.97$, $2.94$, $2.74$, and $2.76$ for the datasets ASSISTments2009, IMDB, MovieLens, and CIFAR-100, respectively.

\begin{figure*}
\centering
  \includegraphics[width=0.9\textwidth]{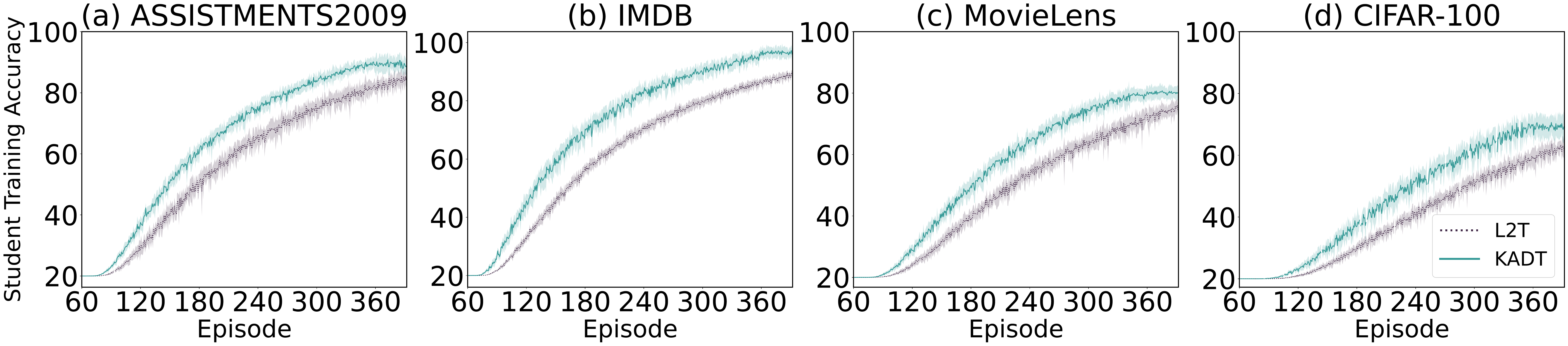}\vspace{-0cm}
   \caption{{Student training accuracy curves averaged over 5 runs, by our KADT method in comparison with the L2T method. (a) DKT student model on the ASSISTments2009 dataset. (b) SVM student model on the IMDB dataset. (c) LSTM student model on the MovieLens dataset. (d) MLP student model on the CIFAR-100 dataset.}}
   \label{fig:teach_beahv_exp}
  \vspace{-0cm}
\end{figure*}

\subsection{Evaluating Knowledge Representation Learning}
\label{sec:ESR}

To evaluate the impact of a student's knowledge representation learning on the effectiveness of a data teaching strategy, we present the progress of a student model's validation accuracy by our KADT method and compare it with the L2T method in Figure~\ref{fig:context}. These heatmaps illustrate the learning progress of a CNN student model on five learning concepts: ``Fish'', ``Insects'', ``Large carnivores'', ``Flowers'', and ``Vehicles'', over the CIFAR-100 dataset.

We observe two main findings from these heatmaps. First, the progress of validation accuracy with the KADT method in the bottom heatmap is more stable and evenly distributed over the five learning concepts than the one with the L2T method in the top heatmap. For example, the final episodes 288-344 over the ``Insects'' and ``Flowers'' learning concepts are more evenly distributed for the KADT method in comparison to the L2T method. Second, the student model with the L2T method suffers from forgetting which can be observed around episodes 144-160 in the ``Fish'' learning concept and around episodes 208-232 in the ``Flowers'' learning concept. 
These findings confirm the effectiveness of learning the performance of a student model on multiple learning concepts in the KADT method, in comparison with learning through a concept-agnostic way in the L2T method.

\subsection{Evaluating Teaching Strategies}
\label{ETB}
In this experiment, we compare the training accuracy of the student models achieved by the KADT and L2T methods, since the L2T method is the state-of-the-art reinforcement learning based teaching model. 

Figure \ref{fig:teach_beahv_exp} shows the average training accuracy curve of DKT \cite{DKT2015} as a student model on the ASSISTments2009 dataset, SVM model on the IMDB dataset, LSTM model on the MovieLens dataset, and MLP model on the CIFAR-100 dataset. We compare the results from the KADT method (green line) with the ones from the L2T method (purple line). It can be observed that the student's training accuracy curves with the KADT method are more stable and converge faster than the ones with the L2T method while achieving a higher training accuracy value after convergence. These findings reflect that a data teaching strategy evolved by the KADT method had a better impact on the student's training performance in comparison to the L2T one.


\begin{table*}
\centering
  \caption{The Average AUC results for comparing different variants of KADT and L2T over all the datasets.}\vspace*{-0cm}
  \label{tbl:ablation}
   \begin{adjustbox}{max width=\textwidth}
  \begin{tabular}{l|ccc|cccc}
    \toprule
     \multirow{2}{*}{Model\hspace*{2cm}}& \multicolumn{3}{c|}{Components}& \multicolumn{4}{c}{Datasets}\\\cline{2-8}
     &RL&KT&Attention&ASSISTments2009& IMDB&MovieLens&CIFAR-100\\
      
      \midrule
     L2T &\checkmark&$\times$&$\times$&$84.31\pm0.03$&$89.46\pm0.06$& $77.23\pm0.03$& $65.33\pm0.06$\\\hline
   KADT-Basic &\checkmark&$\times$&$\times$&$84.69\pm0.01$&$89.83\pm0.03$& $77.70\pm0.04$& $65.87\pm0.03$\\
   KADT-KT&\checkmark&\checkmark&$\times$&$85.79\pm0.02$&$90.84\pm0.04$&$78.90\pm0.03$&$67.11\pm0.02$\\\hline
   KADT&\checkmark&\checkmark&\checkmark&$\mathbf{87.10\pm0.04}$&$\mathbf{92.24\pm0.02}$&$\mathbf{80.31\pm0.02}$&$\mathbf{68.75\pm0.03}$\\
  
   \bottomrule
\end{tabular}
\end{adjustbox}
\end{table*}

\subsection{Ablation Study}
\label{sec:AS}
We conduct an ablation study to assess the effect of each individual component in our KADT method. We consider and compare two different variants including: 1) a baseline variant called ``KADT-Basic'', which follows the same definitions of state and action from the L2T model~\cite{L2T} but uses our reward function design, and 2) a variant with the KT component working with mean pooling, i.e., taking the mean of knowledge representation vectors for samples within the same class label as the knowledge representation for the class label while treating them as equally likely, we call it ``KADT-KT''. Our complete model KADT has both KT and attentive pooling components. We perform multiple independent runs and calculate the statistical significance of the results using student t-test counting a p-value $< 0.05$ as a statistically significant finding.

Table~\ref{tbl:ablation} summarizes the average AUC results for different variants in the ablation study.  We highlight the findings as follows. Firstly, the impact of our reward function design can be observed by comparing the KADT-Basic variant with the L2T model~\cite{L2T}. A significant performance margin of $\mathrm{0.38}$, $\mathrm{0.37}$, $\mathrm{0.47}$, and $\mathrm{0.54}$ exists between these two models for \emph{ASSISTments2009}, \emph{MDB}, \emph{MovieLens}, and \emph{CIFAR-100}, respectively. Secondly, the impact of the KT model can be observed by comparing the KADT-Basic variant with the KADT-KT variant. A statistically significant performance enhancement by a margin of $\mathrm{1.10}$, $\mathrm{1.01}$, $\mathrm{1.20}$, and $\mathrm{1.24}$ is achieved for \emph{ASSISTments2009}, \emph{MovieLens}, and \emph{CIFAR-100}, respectively. Finally, the impact of the attentive pooling technique can be observed by comparing the KADT-KT variant with the complete KADT model. A statistically significant performance enhancement by a margin of $\mathrm{1.31}$, $\mathrm{1.40}$, $\mathrm{1.41}$, and $\mathrm{1.64}$ is achieved for \emph{ASSISTments2009}, \emph{MovieLens}, and \emph{CIFAR-100}, respectively.

%% file: RelatedWork.tex
\section{Related Work}
\label{sec:rl}
In this section, we explore the related work over two themes including machine teaching and learning to teach.

Machine Teaching (MT) is a learning paradigm, commonly referred to as the \emph{inverse problem of machine learning}~\cite{Zhu_iml_15}. It involves interaction between two models including a student model aiming at learning a specific task, and a teacher model targeting to sample optimal training data for the student model. With the rise of deep learning models and the availability of large training data, there is an increasing interest for MT in different application areas such as cyber-security~\cite{Alfeld_2017}, deep learning model compression~\cite{Romero_2015}, or inverse reinforcement learning~\cite{IRL_2019}. MT methods can be categorized into two groups based on the interaction between a student model and a teacher model ~\cite{Liu_icml_17}: \emph{batch machine teaching} and \emph{interactive machine teaching}. 

In batch MT methods, the teaching process is done in an offline manner.
A teacher model learns a ranking function over training data to select a subset of training samples for a student model. Zhu~\cite{Zhu_2013} proposed a Bayesian machine teaching model that can optimize training data by balancing a trade-off between teaching effort and a student's loss. Liu et al.~\cite{Liu_2016} conducted an extensive evaluation for linear student models to find an optimal teaching model that minimizes training data size. There are two main limitations for this group. First, it largely depends on knowledge about the design of student models, which makes it hard to generalize across different student models. Second, it does not consider iterative optimization (e.g., SGD) for student models.

Interactive MT methods work through iterative interactions, where each interaction involves observing the state of a student model followed by sampling a subset of training data. John et al.~\cite{Johns_cvpr_15} introduced an interactive MT model for teaching on image datasets by using a kernel function to rank training samples w.r.t the development of a student's performance. Liu et al.~\cite{liu2017iterative} proposed three approaches for designing a teacher model in an interactive setting: 1) an omniscient teacher that learns to sample a new training example by minimizing the difference (i.e., L2 norm) between its parameters and a student's parameters, 2) a surrogate teacher that learns to minimize the difference between its expected loss and a student's loss given a candidate training sample, and 3) an imitation teacher that learns to imitate a student's hypothesis function to sample a new training example. These methods share two limitations, including dependency on an expert student model which might not exist in some scenarios, and sampling one training example at each interaction, which is less stable during a student's optimization in comparison to sampling a mini-batch of training examples. 

Recently, \emph{Learning to Teach} (L2T)~\cite{L2T} was proposed as a teaching framework that allows customizing the learning process for a student model from three aspects: the selection of training data, the design of a loss function, and the design of a hypothesis function. The authors developed an RL method to optimize a data teaching policy by sampling training mini-batches that help a student model to achieve a target performance threshold \cite{L2T}. Later, Wu et al.~\cite{L2TDF18} studied how to learn a dynamic loss function to better teach a student model based on a gradual optimization method~\cite{HazanLS16}. Albeit, L2T considers only the overall performance of a student model using simple features such as training loss and validation accuracy, while ignoring how performance relates to latent learning concepts. Our proposed KADT method solves this limitation by dynamically tracking the student's performance on multiple learning concepts. Moreover, our method incorporates several novel RL designs to further improve the effectiveness and robustness of learning teaching strategies.

%% file: Conclusion.tex
\section{Conclusion}
\label{sec:conclusion}

In this work, we proposed a novel method KADT which has the ability to dynamically learn knowledge representations of a student model over latent learning concepts during the evolution of a data teaching strategy. Thus, there is no need for manual design or calibration of states across different learning tasks. Further, KADT is developed in a RL framework with several novel design choices, which provides better generalization across different student models and learning tasks. We compared KADT with the state-of-the-art methods. The results showed that KADT consistently outperforms these methods on the four learning tasks. For future work, we will explore other aspects of teaching strategies including loss function and hypothesis function. Moreover, we will explore meta-learning methods for teaching strategy optimization.


\ifCLASSOPTIONcaptionsoff
  \newpage
\fi